\documentclass[lettersize,journal]{IEEEtran}
\usepackage{amsmath,amsfonts,amssymb}
\usepackage{algorithmic}
\usepackage{algorithm}
\usepackage{array}
\usepackage[caption=false,font=normalsize,labelfont=sf,textfont=sf]{subfig}
\usepackage{textcomp}
\usepackage{stfloats}
\usepackage{url}
\usepackage{verbatim}
\usepackage{graphicx}
\usepackage{cite}
\usepackage{booktabs}
\usepackage{multirow}
\usepackage{mathtools}
\usepackage{xcolor}
\usepackage{float}

\begin{document}

\title{Dynamic Traceback Learning for\\ Medical Report Generation}

\author{Shuchang Ye, Mingyuan Meng, Mingjian Li, Dagan Feng, Usman Naseem, and Jinman Kim,~\IEEEmembership{Member,~IEEE}}

\markboth{IEEE Transactions on Multimedia}%
{Shell \MakeLowercase{\textit{et al.}}: A Sample Article Using IEEEtran.cls for IEEE Journals}


\maketitle

\begin{abstract}
Automated medical report generation has demonstrated the potential to significantly reduce the workload associated with time-consuming medical reporting. Recent generative representation learning methods have shown promise in integrating vision and language modalities for medical report generation. However, when trained end-to-end and applied directly to medical image-to-text generation, they face two significant challenges: i) difficulty in accurately capturing subtle yet crucial pathological details, and ii) reliance on both visual and textual inputs during inference, leading to performance degradation in zero-shot inference when only images are available. To address these challenges, this study proposes a novel multimodal dynamic traceback learning framework (\textbf{\textit{DTrace}})\footnote{To facilitate reproducibility, we have made our code publicly available at: \url{https://github.com/ShuchangYe-bib/DTrace}}. Specifically, we introduce a traceback mechanism to supervise the semantic validity of generated content and a dynamic learning strategy to adapt to various proportions of image and text input, enabling text generation without strong reliance on the input from both modalities during inference. The learning of cross-modal knowledge is enhanced by supervising the model to recover masked semantic information from a complementary counterpart. Extensive experiments conducted on two benchmark datasets, IU-Xray and MIMIC-CXR, demonstrate that the proposed \textbf{\textit{DTrace}} framework outperforms state-of-the-art methods for medical report generation.
\end{abstract}


\section{Introduction}
\IEEEPARstart{M}{edical} report generation is a crucial component of the image diagnostic process, providing detailed textual descriptions of medical images to guide clinical decision-making and treatment planning~\cite{survey_rg_2020}. However, the task of interpreting medical images and composing reports is both time-consuming and resource-intensive. Therefore, automating the report generation process to produce a functional draft has garnered significant attention as a potential solution to reduce radiologist workload~\cite{survey_rg_2023}. Despite advances in deep learning, medical report generation remains challenging due to the difficulty in accurately associating subtle yet critical diagnostic features in medical images with their corresponding textual reports. This is because medical images often contain subtle regions, such as tumor areas, which are essential for diagnosis, while medical reports rely on a limited set of keywords to represent this diagnostic information. Hence, it is crucial to develop a method that can capture and associate this nuanced information to preserve the intrinsic medical meanings within medical images and reports.

\begin{figure}[t]
\centering
\includegraphics[width=0.95\columnwidth]{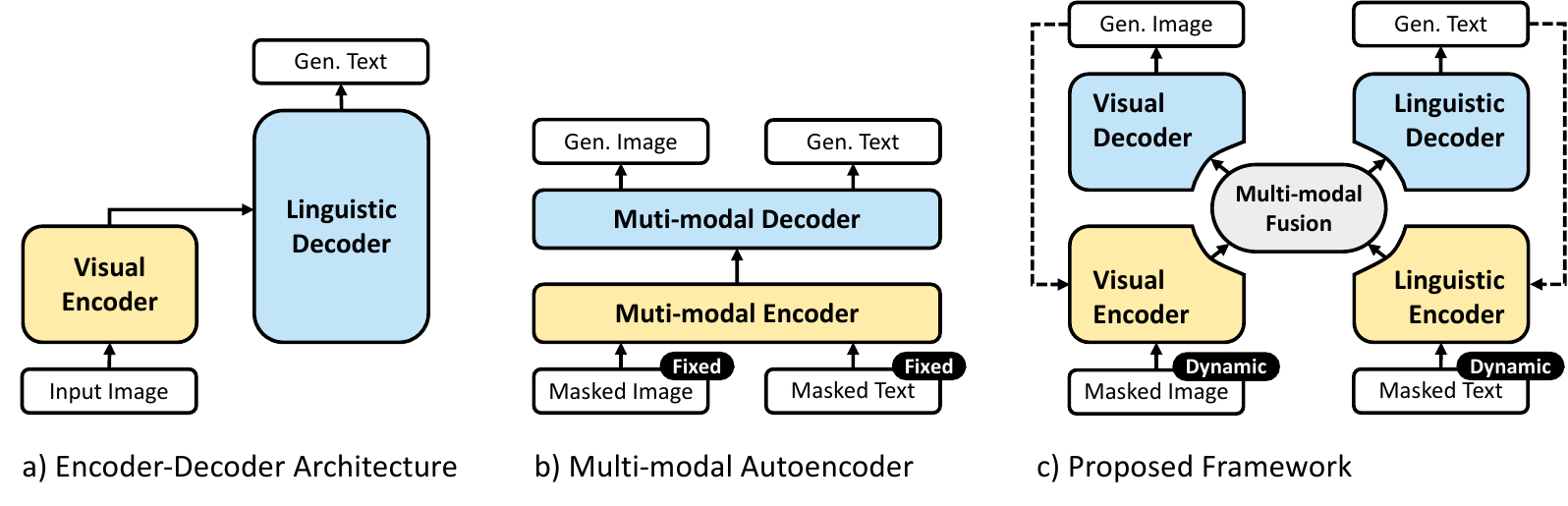} 
\caption{Illustration of different generative frameworks. (a) Common unimodal encoder-decoder framework, (b) Multimodal masked encoder-decoder framework for generative representation learning (GRL), and (c) Our proposed framework with dynamic traceback learning (DTrace).}
\label{gframeworks}
\end{figure}

\begin{figure}[!t]
\centering
\includegraphics[width=0.95\columnwidth]{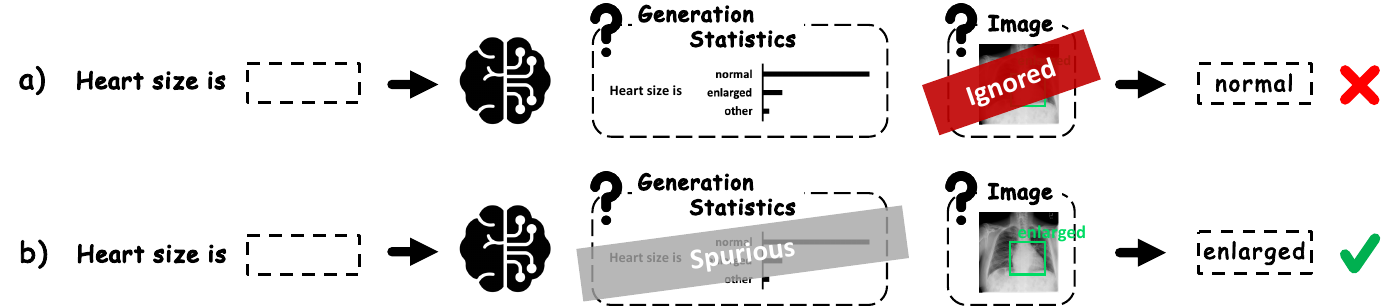} 
\caption{Illustration of the limitation of the existing report generation framework. a) The model makes predictions based on spurious generation statistics, overlooking the radiology images; b) Ideally, the model should understand the pathological information in the image and generate a report accordingly.}
\label{statistics}
\end{figure}

Existing medical report generation methods usually rely on an encoder–decoder framework (Fig.~\ref{gframeworks}a) that performs unimodal learning to build a uni-directional mapping from images to reports~\cite{attention_is_all_you_need,r2gen,pure_transformer,aligntransformer}. While this imposes a forward semantic connection, it neglects the reciprocal relation that textual semantics map back to visual evidence. Without such bi-directional supervision, models often learn to exploit statistical regularities in the text (e.g., frequent phrases or template-like sentences) as shortcuts, rather than grounding their outputs in image evidence~\cite{disease_reveal}, as depicted in Fig.~\ref{statistics}. To overcome this limitation, we argue that text-to-image generation is equally critical. By requiring text embeddings to reconstruct masked image regions, the model is compelled to encode language in a way that preserves pathology-relevant semantics. This bi-directional training enforces stronger cross-modal alignment, ensuring that generated reports are linguistically coherent and semantically grounded in the underlying visual content. Modeling such mutual associations between modalities has been shown to facilitate a more profound understanding of cross-modal knowledge~\cite{multimodal_simulators}. Recently, Generative Representation Learning (GRL) methods have exploited multimodal learning and generation~\cite{bert,mae,m3ae}, where image and report reconstruction are jointly performed in a multimodal masked encoder-decoder framework (Fig.~\ref{gframeworks}b). These methods aim to learn the latent space representations by masking inputs and subsequently reconstructing the original inputs based on the unmasked information. Although these methods have brought significant advancements in multimodal learning, they exhibit considerable gaps when adopted for medical report generation due to two primary limitations:

First, GRL methods prioritize capturing superficial structural cues rather than deeper pathological semantics~\cite{semmae}. In images, this is reflected in morphological properties such as organ contours, shapes, and spatial layouts. In contrast, in reports, it manifests as structural regularities, including the templated phrasing and typical sentence organization of radiology narratives. These modality-specific surface features are often overemphasized, limiting the ability of GRL methods to represent fine-grained pathological semantics, such as lesion extent or disease progression. In image reconstruction, these methods often reduce the task to simple pixel matching by minimizing pixel-level discrepancies, neglecting the semantic and pathological nuances of images. For report reconstruction, GRL methods tend to predict frequently observed words to achieve a high overlap rate between the original and reconstructed reports. Due to the word imbalance in medical reports, where keywords signifying pathology appear infrequently, this learning approach potentially leads to the generation of clinically-flawed templated reports. Further, inherent variability exists in the report descriptions for the same medical images, such as the reporting order (e.g., starting from the disease sites or from image acquisition protocol) and the lexical choices employed to convey the severity of symptoms. However, the loss functions of GRL methods typically operate at the word level, which cannot measure the quality of generated reports from the sentence level with semantic contextual information.

Second, GRL methods focus on predicting masked images and text from their unmasked counterparts, which usually require a large amount of unmasked information to achieve text reconstruction~\cite{bert,m3ae}, with the mask ratio for the text being limited to a low range, $\approx 15\%$ (most textual information retained as the input). Meanwhile, the inference is based solely on images in medical report generation. Such a shift can lead to a performance drop (see Section~\ref{impact_of_mask_ratios}), as GRL methods are not designed to handle inference from images alone without accompanying text.

To address the above limitations, in this study, we propose a novel report generation framework to overcome the limitations of GRL methods and introduce dynamic traceback learning for medical report generation (DTrace). During inference, images are fully visible ($0\%$ mask ratio), and text is completely masked ($100\%$ mask ratio). During training, image and text mask ratios fluctuate between $0\%$ and $100\%$  per batch. We demonstrate that this variable masking approach is the most effective way of learning (see Section~\ref{impact_of_mask_ratios}). Our contributions are as follows:

\begin{itemize}
    \item We introduce DTrace, a multimodal framework for medical report generation that jointly learns bi-directional image-to-report and report-to-image generation, enhancing cross-modal knowledge by recovering masked semantic information.
    \item We introduce a traceback mechanism in DTrace that ensures semantic validity by reintegrating generated images and reports into their encoders for self-assessment.
    \item We introduce a dynamic learning strategy in DTrace that adapts to any image-text ratio, enabling effective training with both modalities and supporting image-only inference by adjusting loss weights dynamically.
    \item Extensive experiments on two benchmark datasets (IU-Xray and MIMIC-CXR) show that the proposed DTrace outperforms state-of-the-art medical report generation methods.
\end{itemize}

\section{Related Work}

\subsection{Medical Report Generation} 

Traditional medical report generation methods rely on rule- or template-based methods~\cite{traditional_report_generation}. Rule-based methods often fall short in handling different scenarios and capturing language subtleties, while template-based methods are dependent on template quality and adaptability. With the paradigm shift in Computer Vision (CV) and Natural Language Processing (NLP), deep learning-based medical report generation methods have achieved promising performance and attained wide attention~\cite{survey_rg_2020}.

Deep learning–based report generation can be traced back to the encoder–decoder architecture, where images are transmuted into representative vectors through a visual encoder and then decoded into text~\cite{show_tell}. Subsequent studies enhanced both components~\cite{from_show_to_tell}, evolving from CNN–RNN to Vision Transformer and Transformer~\cite{pure_transformer}, and later turned to improving interaction between the encoder and decoder. Visual language pretraining (VLP) has further advanced image-to-text generation by integrating visual and textual data; for example, MCGN~\cite{mcgn} employs CLIP~\cite{clip} with contrastive learning to refine latent spaces and boost semantic similarity. However, VLP approaches mainly strengthen encoders, while report generation also requires effective decoder training. In medical report generation, R2GenCMN~\cite{r2gen_cmn} unified visual and linguistic representations with a shared vector pool, and XProNet~\cite{xpronet} improved it using pathological labels, but both still relied on uni-directional image-to-report mapping. Recent methods incorporate additional annotations: COMG~\cite{comg} uses segmentation masks to guide attention to critical regions, and RGRG~\cite{rgrg} links bounding boxes with sentences to focus on specific anatomy. Others emphasize semantic alignment, such as ICT~\cite{ict}, which proposes a dual-level collaboration and inference network to enhance sentence coherence and report-level fidelity.

\subsection{Generative Representation Learning}

GRL methods learn the latent space representations by training the model to reconstruct the masked inputs based on the unmasked information. Masked Image Modeling (MAE)~\cite{mae} and Masked Language Modeling (BERT)~\cite{bert} are prevalent pretraining techniques in CV and NLP. BERT learned the word latent representations by training the model to predict the masked content based on the surrounding words. Then, MAE employed a similar strategy in images, where the images were split into patches and the model was trained to reconstruct the randomly masked patches. Recently, M3AE~\cite{m3ae} exploited multimodal learning and generation, where image and text reconstruction were jointly performed with a multimodal masked encoder-decoder framework to enhance the comprehension of cross-modal associations. These GRL methods were widely adopted as a pretraining step to enhance the performance of downstream tasks, such as disease classification~\cite{disease_classification} and medical visual question answering~\cite{question_answering}. 
Subsequently, the concept of masking-based generative representation learning has been extended to various domains, including video understanding, as demonstrated by methods like MGMAE~\cite{mgmae}, and the enhancement of foundation model training, as exemplified by approaches such as EfficientSAM~\cite{efficientsam}.

The capability of reconstructing masked text provides the potential for report generation. However, GRL methods were seldom applied to medical report generation due to the two drawbacks that we identified above. Recently, MedViLL~\cite{medvill} applied the GRL framework to medical report generation (as one of the downstream tasks) by progressively replacing mask tokens with predicted language tokens. Unfortunately, its performance was limited (0.066 in BLEU4 in the MIMIC-CXR dataset) as it was not trained to handle situations where text information is not available.

In contrast, our proposed framework differs in several key aspects: i) GRL methods are typically designed for pre-training on large datasets followed by fine-tuning for specific tasks, whereas our framework is trained end-to-end and can be directly applied to medical report generation; ii) GRL usually employs a fixed mask ratio, but we utilize a dynamic mask ratio; and iii) while conventional methods prioritize morphological and structural similarity, our approach emphasizes semantic similarity.

\begin{figure*}[t]
\centering
\includegraphics[width=0.95\linewidth]{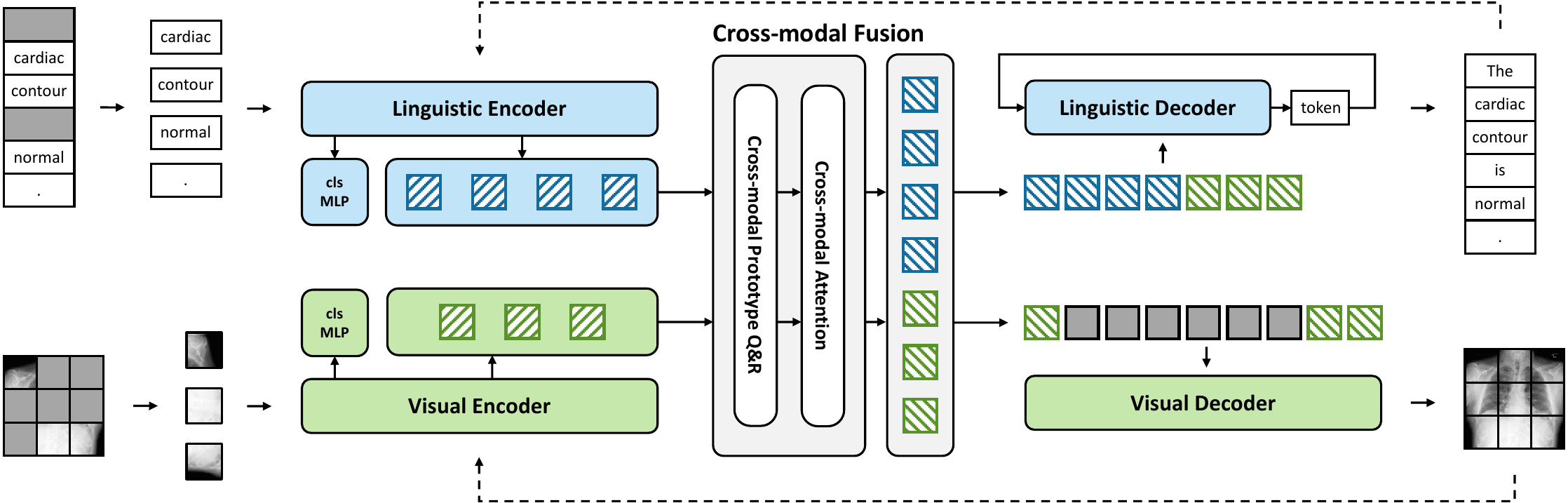} 
\caption{DTrace with dynamic traceback learning. Solid and dashed lines indicate forward and traceback stages. Training: (1) Forward stage: masked images and masked reports are encoded by the visual and linguistic encoders, fused by the cross-modal module, and decoded to reconstruct images and reports. (2) Traceback stage: the reconstructed images and reports are re-encoded to verify semantic validity, with losses computed against unmasked ground-truth. Inference: the visual encoder processes the full image, and at each decoding step, the linguistic encoder encodes the autoregressively generated prefix tokens; fused features are passed to the linguistic decoder to predict the next report token until an end token is generated.}
\label{dtrace}
\end{figure*}

\section{Method}

\subsection{Network Architecture}

DTrace consists of five key components: a visual encoder, a visual decoder, a linguistic encoder, a linguistic decoder, and a cross-modal fusion module (Fig.~\ref{dtrace}). The visual encoder processes partially masked images to extract incomplete pathological information. Simultaneously, the linguistic encoder processes fragmented textual information to extract incomplete pathological information. Then, the extracted information is fed into the cross-modal fusion module, where a cross-attention mechanism is employed to foster the interaction between the visual and linguistic domains, thereby ameliorating the semantic deficits in both modalities. After this, the enriched information is conveyed to the visual and linguistic decoders to restore the masked images and reports to their original unmasked states. Below, we discuss the key components of DTrace in detail:

\textbf{Visual Encoder}: consistent with the common practice of MAE; medical images were split into patches and then randomly masked. The visual encoder was a standard ViT~\cite{vit}, which mapped the unmasked patches into latent representations and then performed multi-label classification to predict the disease labels extracted via CheXbert~\cite{chexbert}. 

\textbf{Linguistic Encoder}: medical reports were mapped into embedded text tokens following the standard pre-processing steps~\cite{bert}. Then, these tokens were randomly masked and fed into the linguistic encoder. The linguistic encoder was a classic text transformer block~\cite{attention_is_all_you_need}. 

\textbf{Cross-Modal Fusion Module}: the features extracted from both the encoders were projected to a pre-defined dimension, which was then concatenated and fed to a cross-modal attention module~\cite{m3ae} for information interchange. The resultant features were subsequently separated and mapped into latent representations via Multi-layer Perceptrons (MLPs)~\cite{mlp}.  

\textbf{Visual Decoder}: the masked tokens were reinstated to their original positions, aligning with the unmasked encoding tokens. Following this, a lite version of ViT~\cite{mae} was used to restore the masked patches.  

\textbf{Linguistic Decoder}: a relational memory module is incorporated to capture recurrent report patterns across similar images. At each decoding step $t$, the memory matrix from the previous step ($M_{t-1}$) serves as the query, while its concatenation with the previous output embedding $y_{t-1}$ serves as the key and value for the multi-head attention. The updated memory $M_t$ is further refined through residual connections and gated mechanisms to stabilize gradient propagation.

\subsection{Traceback Mechanism}

\begin{figure}[b]
\centering
\includegraphics[width=0.95\columnwidth]{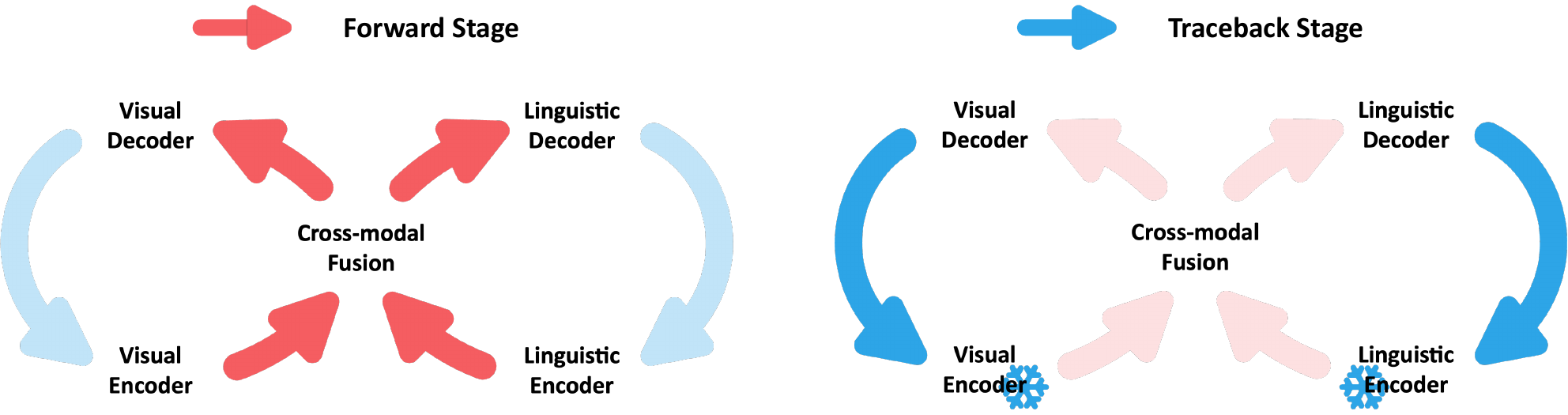} 
\caption{Dynamic traceback learning with forward stage (left) and traceback stage (right). The freeze means that there is no gradient descent in back-propagation.}
\label{phases}
\end{figure}

The traceback mechanism (Fig.~\ref{phases}) was developed to ensure the medical validity of generated outputs through two phases: forward and traceback. Initially, encoder capabilities are enhanced to identify pathological information during the forward phase. Subsequently, the decoder outputs are redirected to their corresponding encoders in the trackback stage to check their medical validity. Through repetitive iterations between the two stages, the DTrace model enhances accuracy by refining its understanding of medical content, ensuring reliable identification and validation of pathological information in image reconstruction and report generation.

\begin{algorithm}[t]
\caption{Dynamic Traceback Learning}
\label{alg:algorithm}
\begin{algorithmic}[1] 
\STATE Generate random numbers: $\alpha$ as image mask ratio and $\beta = 1 - \alpha$ as text mask ratio
\WHILE{Not reach max epoch and early stop}
\IF {Forward}
\STATE imasked, tmasked $\leftarrow$ masking(image, text)
\STATE ifeats, tfeats $\leftarrow$ encoders(imasked, tmasked)
\STATE compute $\ell_{\text{\textit{FVD}}}$ and $\ell_{\text{\textit{FLD}}}$
\STATE gradient descent encoders by $(1-\alpha) \cdot \ell_{\text{\textit{FVD}}}$ and $(1-\beta) \cdot \ell_{\text{\textit{FLD}}}$
\STATE feats $\leftarrow$ cross modal fusion(ifeats, tfeats)
\STATE igen, tgen $\leftarrow$ decoders(feats)
\STATE compute $\ell_{\text{\textit{IR}}}$ and $\ell_{\text{\textit{RG}}}$
\STATE gradient descent all components by $\alpha \cdot \ell_{\text{\textit{IR}}}$ and $\beta \cdot \ell_{\text{\textit{RG}}}$
\ENDIF
\IF {Traceback}
\STATE disable the gradient descent of encoders
\STATE igfeats, tgfeats $\leftarrow$ encoders(igen, tgen)
\STATE compute $\ell_{\text{\textit{TVD}}}$ and $\ell_{\text{\textit{TLD}}}$
\STATE gradient descent all except encoders by $\alpha \cdot 
e^{-\ell_{\text{\textit{FVD}}}} \ell_{\text{\textit{TVD}}}$ and $\beta \cdot 
e^{-\ell_{\text{\textit{FLD}}}} \ell_{\text{\textit{TLD}}}$
\ENDIF
\ENDWHILE
\end{algorithmic}
\end{algorithm}

\subsection{Dynamic Learning Strategy}

Existing multimodal GRL methods incurred performance degradation when generating reports from images alone. To address this limitation, the dynamic learning strategy was proposed to enhance the generalizability of the model to perform multimodal generation given any percentage of text and image inputs. This was achieved through training with various complementary image and text mask ratios (different mask ratios for each batch). This complementary relation aimed to guarantee sufficient shared information and to ensure that each modality can consistently derive some information from the other. It also incorporated a self-adjustment mechanism for loss weights, dynamically adapting to changes in the mask ratios. The pseudo-code of the dynamic traceback learning is shown in Algorithm~\ref{alg:algorithm}. In the subsequent sections, the mathematical rationale behind the adjustment of loss weights during the forward and traceback stages is presented.

Assumed the mask ratio of images to be a random number $0 \le \alpha \le 1$. The corresponding mask ratio of reports would then be $\beta = 1 - \alpha$. Notably, when images are heavily masked, retaining informative textual tokens becomes crucial, as they provide semantic anchors that enrich morphological recovery with diagnostic meaning; conversely, when text is sparsely available, visual morphology compensates, ensuring balanced cross-modal fusion. Although complementary mask ratios do not strictly guarantee non-overlapping information, the stochastic token- and patch-level masking in DTrace ensures that, across training iterations, both modalities encounter diverse missing-information scenarios. This randomness, when combined with semantic supervision from the traceback stage, encourages robust cross-modal compensation even in the presence of partial redundancy~\cite{bert, mae, m3ae}, thereby enabling effective information transfer between modalities.

\textbf{Forward Stage} The forward stage of the DTrace model performs three main tasks: 1) disease identification by the encoders, 2) image reconstruction by the visual decoder, and 3) report generation by the linguistic decoder, as shown on the left of Fig.~\ref{phases}.

For disease identification, we integrated a multi-label classification head for $N$ classes into the encoders to predict disease labels as output. According to the dynamic learning strategy, the image and reports are masked in the forward stage based on the dynamic complementary mask ratio. Subsequently, the remaining unmasked components are directed to their respective encoders. To optimize the generation of their corresponding label outputs, both the encoders aim to minimize the diagnostic loss. The diagnostic loss is the binary cross-entropy (\textit{BCE}) loss (see Equation~\ref{bce_loss}) between the instances of disease label extracted by CheXbert, $y$, and the encoders' corresponding predicted label, $\hat{y}$. As a diagnosis can only be meaningful when sufficient unmasked information is available, the visual and linguistic encoders' respective diagnostic losses complement their corresponding image and text mask ratios. The resulting forward visual diagnostic (\textit{FVD}) loss and forward linguistic diagnostic (\textit{FLD}) loss are shown in Equation \ref{fvd_loss} and Equation \ref{fld_loss}.

\begin{equation}
    \label{bce_loss}
    \ell_{\text{\textit{BCE}}}(y, \hat{y}) = -\frac{1}{N} \sum_{i=1}^{N}y_i \cdot log(\hat{y}_i)+(1-y_i) \cdot log(1-\hat{y}_i)
\end{equation}

\begin{equation}
    \label{fvd_loss}
    \ell_{\text{\textit{FVD}}}(y, \hat{y}) = (1-\alpha) \cdot \ell_{\text{\textit{BCE}}}(y, \hat{y})
\end{equation}

\begin{equation}
    \label{fld_loss}
    \ell_{\text{\textit{FLD}}}(y, \hat{y}) = (1-\beta) \cdot \ell_{\text{\textit{BCE}}}(y, \hat{y})
\end{equation}

For image reconstruction, we follow the MAE's practice~\cite{mae} in that the visual decoder takes the visual features that gained knowledge from the cross-model fusion module and the learned vector that indicates the presence of the masked patch as input. The image reconstruction process is regulated through minimizing the image reconstruction (\textit{IR}) loss, which is the pixel-wise mean-square-error (\textit{MSE}) weighted proportional to the image mask ratio to facilitate cross-modal communication, as shown in Equation~\ref{ir_loss}. In Equation~\ref{ir_loss}, $g$ represents an instance of the origin image and $\hat{g}$ represents the respective reconstructed image, and $W$ and $H$ represent the width and height of the image, respectively.

\begin{equation}
    \label{ir_loss}
    \ell_{\text{\textit{IR}}}(g, \hat{g}) = \alpha \cdot \frac{1}{W} \frac{1}{H} \sum_{i=1}^{W} \sum_{j=1}^{H} (g_{ij}-\hat{g}_{ij})^2
\end{equation}

The generated reports are refined by minimizing the report generation (\textit{RG}) loss. The \textit{RG} loss, as shown in Equation~\ref{rg_loss}, is the word-level cross-entropy (CE) loss weighted corresponding to the text mask ratio to attend to instances where reports are generated solely from images, where $r$ represents an instance of ground truth report, $\hat{r}$  represents the corresponding generated report, $V$ represents the vocabulary size and $L$ represents the report length.

\begin{equation}
    \label{rg_loss}
    \ell_{\text{\textit{RG}}}(r, \hat{r}) = \beta \cdot \frac{1}{V} \frac{1}{L} \sum_{i=1}^{V} \sum_{j=1}^{L} r_{ij} \cdot log(\frac{e^{\hat{r}_{ij}}}{\sum_{k=1}^{V}e^{\hat{r}_{ik}}})
\end{equation}

\textbf{Traceback Stage} As depicted on the right of Fig.~\ref{phases}, the images reconstructed and reports generated by the decoders were traced back to be the input of encoders with frozen parameters. The encoders inference on both the reconstructed items ($\hat{\hat{y}}$) and unmasked items ($\tilde{y}$). To avoid error accumulation when reconstructed outputs are fed back into the encoders, traceback losses are inversely scaled by forward errors, such that unreliable reconstructions are down-weighted while accurate outputs progressively gain influence, ensuring stable and semantically valid feedback, as shown in Equation \ref{tvd_loss} and Equation \ref{tld_loss}. As the forward errors decrease and reconstructions become more accurate, the traceback losses naturally gain more weight, thereby progressively reinforcing semantic consistency without compounding errors.

\begin{equation}
    \label{tvd_loss}
    \ell_{\text{\textit{TVD}}}(\tilde{y}, \hat{\hat{y}}) = \alpha \cdot e^{-\ell_{\text{\textit{FVD}}}(\tilde{y}, \hat{\hat{y}})} \cdot \ell_{\text{\textit{BCE}}}(\tilde{y}, \hat{\hat{y}})
\end{equation}

\begin{equation}
    \label{tld_loss}
    \ell_{\text{\textit{TLD}}}(\tilde{y}, \hat{\hat{y}}) = \alpha \cdot e^{-\ell_{\text{\textit{FLD}}}(\tilde{y}, \hat{\hat{y}})} \cdot \ell_{\text{\textit{BCE}}}(\tilde{y}, \hat{\hat{y}})
\end{equation}

\section{Experimental Setup}

\subsection{Dataset}

We conducted experiments on two well-benchmarked public datasets: Indiana University Chest X-ray (IU-Xray)~\cite{iu_xray} and MIMIC Chest X-ray (MIMIC-CXR)~\cite{mimic_cxr}. We split the data into training, validation, and testing subsets. For IU-Xray, we adopt the widely accepted 7:1:2 data split as suggested in prior studies~\cite{r2gen,r2gen_cmn,xpronet}. For MIMIC-CXR, we adhere to the official data split.

\subsection{Evaluation Metrics}
To measure the quality of generated medical reports, we follow the standard practice~\cite{pure_transformer,r2gen,r2gen_cmn,xpronet} to adopt natural language generation (NLG) metrics: BLEU~\cite{bleu}, METEOR~\cite{meteor}, ROUGE-L~\cite{rouge}, and CIDEr~\cite{cider} as the evaluation metrics. We further assess the clinical efficacy (CE)~\cite{r2gen,r2gen_cmn, organ,kiut} of the generated reports by annotating them with CheXbert and comparing the predicted and ground truth labels. 

\subsection{Implementation Details}\label{imdetails}

Our implementation of DTrace was realized using the PyTorch package~\cite{pytorch}. Optimization of the gradient descent process was carried out utilizing the AdamW optimizer~\cite{AdamW}, with a set learning rate of $10^{-4}$. In the context of report generation, the beam search algorithm was employed, with a specified beam width of $3$. The predefined maximum lengths for the reports were set at $60$ and $100$ for the IU-Xray and MIMIC-CXR datasets, respectively. Training of the model was performed on an NVIDIA RTX A6000 graphics card, with a designated mini-batch size of $16$. To enhance the efficiency of training, the model initially underwent a pre-training phase, wherein it was conditioned to reconstruct images and reports based on the information inherent to their respective modalities, coupled with contrastive learning techniques.

All reported results and discussions in this manuscript are based on an image-only inference setting, in which all text inputs are fully masked during testing to rigorously evaluate the model’s ability to generate reports solely from image inputs. In terms of real-world deployment, the deployed DTrace model has a size of $1,324.74$ MB, an average inference latency of $755.77$ ms per batch, and an inference throughput of $1.32$ FPS.

\begin{table*}[!t]
\centering
\caption{Performance comparison between DTrace and existing report generation methods on the IU-Xray and MIMIC-CXR datasets. The best results are highlighted in bold, and the second best are underlined. BL, MTR, RG-L, and CDr are the abbreviations of NLG evaluation metrics BLEU, METEOR, ROUGE-L, and CIDEr. P, R, and F are the abbreviations of CE metrics: Precision, Recall, and F1-score. Gray indicates the utilization of additional annotations. * denotes that the results are cited from their original papers.}
\label{sotas_comparison}
\begin{tabular}{l|l|ccccccc|ccc}
    \toprule
    Dataset & Model & BL-1 & BL-2 & BL-3 & BL-4 & MTR & RG-L & CDr & P & R & F \\
    \midrule
    \multirow{17}{*}{IU-Xray} 
    & R2Gen & 0.470 & 0.304 & 0.211 & 0.157 & 0.197 & 0.364 & 0.342 & - & - & - \\
    & R2GenCMN & 0.486 & 0.307 & 0.216 & 0.156 & 0.212 & 0.374 & 0.331 & - & - & - \\
    & CMCL* & 0.473 & 0.305 & 0.217 & 0.162 & 0.186 & 0.378 & - & - & - & - \\
    & AlignTransformer* & 0.484 & 0.313 & 0.225 & 0.173 & 0.204 & 0.379 & - & - & - & - \\
    & TriNet* & 0.478 & 0.344 & 0.248 & 0.180 & - & 0.398 & 0.439 & - & - & - \\
    & MCTransformer* & 0.496 & 0.319 & 0.241 & 0.175 & - & 0.377 & \underline{0.449} & - & - & - \\
    & M2KT* & 0.497 & 0.319 & 0.230 & 0.174 & - & 0.399 & 0.407 & - & - & - \\
    & ORGAN* & 0.494 & 0.335 & 0.247 & \underline{0.190} & 0.203 & \underline{0.395} & - & - & - & - \\
    & KiUT & \textbf{0.525} & \textbf{0.360} & \underline{0.251} & 0.185 & \textbf{0.242} & \textbf{0.409} & - & - & - & - \\
    & RAMT & 0.482 & 0.310 & 0.221 & 0.165 & 0.195 & 0.377 & - & - & - & - \\
    & Pretrain+FMVP & 0.485 & 0.315 & 0.225 & 0.169 & 0.201 & 0.398 & - & - & - & - \\
    & MedLLM & - & - & - & 0.168 & 0.209 & 0.381 & 0.427 & - & - & - \\
    & MultiP & 0.515 & 0.332 & 0.238 & 0.178 & 0.216 & 0.386 & - & - & - & - \\
    & CK-Net & 0.503 & 0.325 & 0.236 & 0.179 & 0.202 & 0.373 & - & - & - & - \\
    & DTrace (ours) & \underline{0.516} & \underline{0.353} & \textbf{0.278} & \textbf{0.204} & \underline{0.233} & \underline{0.386} & \textbf{0.469} & - & - & - \\
    & \textcolor{gray}{Oracle+FMVP*} & \textcolor{gray}{0.499} & \textcolor{gray}{0.339} & \textcolor{gray}{0.256} & \textcolor{gray}{0.206} & \textcolor{gray}{0.211} & \textcolor{gray}{0.422} & \textcolor{gray}{-} & \textcolor{gray}{-} & \textcolor{gray}{-} & \textcolor{gray}{-} \\
    & \textcolor{gray}{COMG*} & \textcolor{gray}{0.482} & \textcolor{gray}{0.316} & \textcolor{gray}{0.233} & \textcolor{gray}{0.184} & \textcolor{gray}{0.191} & \textcolor{gray}{0.382} & \textcolor{gray}{-} & \textcolor{gray}{-} & \textcolor{gray}{-} & \textcolor{gray}{-} \\
    \midrule
    \multirow{17}{*}{MIMIC-CXR}
    & R2Gen & 0.344 & 0.208 & 0.140 & 0.100 & 0.135 & 0.271 & 0.146 & 0.333 & 0.273 & 0.276 \\
    & R2GenCMN & 0.327 & 0.211 & 0.148 & 0.109 & 0.137 & \underline{0.298} & 0.135 & 0.334 & 0.275 & 0.278 \\
    & CMCL* & 0.344 & 0.217 & 0.140 & 0.097 & 0.133 & 0.281 & - & - & - & - \\
    & AlignTransformer* & 0.378 & 0.235 & 0.156 & 0.112 & 0.158 & 0.283 & - & - & - & - \\
    & TriNet* & 0.362 & 0.251 & 0.188 & 0.143 & - & 0.326 & 0.273 & - & - & - \\
    & MCTransformer* & 0.351 & 0.223 & 0.157 & 0.118 & - & 0.287 & \underline{0.281} & - & - & - \\
    & M2KT* & 0.386 & 0.237 & 0.157 & 0.111 & - & 0.274 & 0.111 & - & - & - \\
    & ORGAN* & \textbf{0.405} & \underline{0.254} & \underline{0.170} & \underline{0.121} & \underline{0.161} & 0.291 & - & 0.416 & \underline{0.418} & \underline{0.385} \\
    & KiUT & 0.393 & 0.243 & 0.159 & 0.113 & 0.160 & 0.285 & - & 0.371 & 0.318 & 0.321 \\
    & RAMT & 0.362 & 0.229 & 0.157 & 0.113 & 0.153 & 0.284 & - & 0.380 & 0.342 & 0.335 \\
    & Pretrain+FMVP & 0.389 & 0.236 & 0.156 & 0.108 & 0.150 & 0.284 & - & - & - & - \\
    & MedLLM & - & - & - & 0.128 & 0.161 & 0.289 & 0.265 & 0.294 & 0.277 & 0.281 \\
    & CK-Net & 0.374 & 0.235 & 0.157 & 0.116 & 0.150 & 0.284 & - & 0.386 & 0.379 & 0.384 \\
    & DTrace (ours) & \underline{0.392} & \textbf{0.260} & \textbf{0.171} & \textbf{0.129} & \textbf{0.162} & \textbf{0.309} & \textbf{0.311} & \underline{0.411} & \textbf{0.436} & \textbf{0.391} \\
    & \textcolor{gray}{Oracle+FMVP*} & \textcolor{gray}{0.391} & \textcolor{gray}{0.249} & \textcolor{gray}{0.172} & \textcolor{gray}{0.125} & \textcolor{gray}{0.160} & \textcolor{gray}{0.304} & \textcolor{gray}{-} & \textcolor{gray}{-} & \textcolor{gray}{-} & \textcolor{gray}{-} \\
    & \textcolor{gray}{COMG*} & \textcolor{gray}{0.346} & \textcolor{gray}{0.216} & \textcolor{gray}{0.145} & \textcolor{gray}{0.104} & \textcolor{gray}{0.137} & \textcolor{gray}{0.279} & \textcolor{gray}{-} & \textcolor{gray}{0.424} & \textcolor{gray}{0.291} & \textcolor{gray}{0.345} \\
    & \textcolor{gray}{RGRG} & \textcolor{gray}{0.373} & \textcolor{gray}{0.249} & \textcolor{gray}{0.175} & \textcolor{gray}{0.126} & \textcolor{gray}{0.168} & \textcolor{gray}{0.264} & \textcolor{gray}{0.495} & \textcolor{gray}{0.461} & \textcolor{gray}{0.475} & \textcolor{gray}{0.447} \\
    \bottomrule    
\end{tabular}
\end{table*}

\subsection{Baselines}

We compared the performance of DTrace against the state-of-the-art medical report generation methods (see Section~\ref{sec:comparison}): R2Gen~\cite{r2gen}, R2GenCMN~\cite{r2gen_cmn}, CMCL~\cite{cmcl}, AlignTransformer~\cite{aligntransformer}, TriNet~\cite{trinet}, MCTransformer~\cite{pure_transformer}, M2KT~\cite{m2kt}, ORGAN~\cite{organ}, KiUT~\cite{kiut}, RAMT~\cite{ramt}, pretrain+FMVP~\cite{fmvp}, Med-LLM~\cite{medllm}, MultiP~\cite{multip} and CK-Net~\cite{cknet}. For fair evaluation, the released code from the baseline methods was used in the same settings as described in the papers. Additionally, we included methods such as COMG~\cite{comg}, oracle+FMVP~\cite{fmvp}, and RGRG~\cite{rgrg}, which leverage auxiliary annotations beyond textual reports, to provide a holistic benchmarking comparison. To further analyze DTrace's capabilities, we performed an ablation study to evaluate the contributions of individual components. To quantify the traceback paradigm's unique advantages against reconstruction-based approaches, we compare our Dtrace against M3AE~\cite{m3ae}, which reconstructs masked image patches and report tokens using standard reconstruction losses only, without applying traceback or weighted diagnostic supervision. We also investigated the impact of varying the mask ratio on model performance during training. Finally, we present qualitative comparisons between the reports generated by DTrace and those produced by competing methods, offering insights into the semantic and contextual differences.

\section{Result and Discussion}

\subsection{Comparisons to Previous Methods}
\label{sec:comparison}

Table~\ref{sotas_comparison} presents a comprehensive comparison with state-of-the-art methods, highlighting the superior performance of the proposed DTrace across both NLG and clinical efficacy (CE) metrics on the IU-Xray and MIMIC-CXR datasets. DTrace consistently ranks among the top performers in NLG metrics, particularly excelling in BLEU-3, BLEU-4, and CIDEr, which underscores its strong capability in generating structurally accurate and linguistically coherent reports. In terms of CE metrics, DTrace achieves the best results in recall and F1-score and the second-best in precision, demonstrating its effectiveness in capturing disease regions and generating semantically meaningful reports. Even in comparison with models like COMG and RGRG, which leverage additional segmentation masks and region-specific information, DTrace remains highly competitive.

This superior performance suggests that DTrace is adept at capturing pathology-critical information, making it well-suited for scenarios where radiologists' descriptions of the same radiology image vary in writing style and terminology. For example, when comparing the ground truth sentence ``the heart size is normal" with two variants, ``the heart size is enlarged" and ``the heart size is within normal limits," the cross-entropy is smaller for the first variant. To address this potential bias towards common phrasing, which might sacrifice semantic accuracy, we introduce a traceback mechanism. This mechanism evaluates generated reports using an encoder trained for accurate diagnosis, thus reinforcing semantic correctness and diluting the impact of cross-entropy loss.

The proposed DTrace outperforms models such as R2GenCMN, AlighTransformer, and XProNet, which we attribute to the benefits of multimodal learning, particularly in facilitating cross-modal communication. The traceback mechanism further contributes to this performance improvement. Even when a modality lacks information from another, it can reconstruct its form independently, while the traceback mechanism ensures that semantic information is still obtained from another modality. In the cross-modal fusion module, different modalities exchange and complement each other's information, thereby establishing a robust communication protocol. This approach achieves effects similar to R2GenCMN and XProNet, with the added enhancement of the traceback mechanism, leading to overall improved performance.

\begin{table*}[t]
\caption{Ablation study on key components of our proposed DTrace on the MIMIC-CXR dataset.}
\label{ablation_study}
\centering
\begin{tabular}{l|ccccccc|ccc}
    \toprule
    Model & BL-1 & BL-2 & BL-3 & BL-4 & MTR & RG-L & CDr & P & R & F \\
    \midrule
    \textcolor{gray}{Unimodal Report Auto-completion} & \textcolor{gray}{0.351} & \textcolor{gray}{0.215} & \textcolor{gray}{0.141} & \textcolor{gray}{0.102} & \textcolor{gray}{0.133} & \textcolor{gray}{0.264} & \textcolor{gray}{0.108} & \textcolor{gray}{0.277} & \textcolor{gray}{0.244} & \textcolor{gray}{0.236} \\ 
    \midrule
    Encoder-Decoder (Baseline) & 0.348 & 0.212 & 0.143 & 0.106 & 0.136 & 0.277 & 0.143 & 0.325 & 0.271 & 0.268 \\
    + Bi-directional Generation & 0.346 & 0.220 & 0.144 & 0.107 & 0.142 & 0.285 & 0.241 & 0.345 & 0.280 & 0.280 \\
    \, + Dynamic Learning & 0.371 & 0.243 & 0.165 & 0.120 & 0.155 & 0.281 & 0.279 & 0.358 & 0.355 & 0.344 \\
    \, \, + Traceback Mechanism & 0.392 & 0.260 & 0.171 & 0.129 & 0.162 & 0.309 & 0.311 & 0.411 & 0.436 & 0.391 \\
    \midrule
    \textcolor{gray}{Multimodal Masked Autoencoder} & \textcolor{gray}{0.364} & \textcolor{gray}{0.246} & \textcolor{gray}{0.164} & \textcolor{gray}{0.119} & \textcolor{gray}{0.153} & \textcolor{gray}{0.284} & \textcolor{gray}{0.292} & \textcolor{gray}{0.354} & \textcolor{gray}{0.296} & \textcolor{gray}{0.301} \\ 
    \bottomrule    
\end{tabular}
\end{table*}

\subsection{Ablation Study}
\label{sec:ablation_study}

\begin{table*}[t]
\centering
\caption{Comparison of the traditional fixed mask ratio training strategy versus our dynamic learning strategy on the IU-Xray dataset, with our method highlighted in bold.}
\label{masks_comparison}
\begin{tabular}{lcccccccc}
    \toprule
    Strategy & Mask Ratio & BL-1 & BL-2 & BL-3 & BL-4 & MTR & RG-L & CDr \\
    \midrule
    \multirow{7}{*}{Fixed} 
    & 0\% & 0.397 & 0.255 & 0.183 & 0.139 & 0.164 & 0.367 & 0.364 \\
    & 15\% & 0.441 & 0.284 & 0.207 & 0.158 & 0.180 & 0.378 & 0.427 \\
    & 30\% & 0.459 & 0.293 & 0.207 & 0.155 & 0.176 & 0.368 & 0.255 \\
    & 45\% & 0.465 & 0.299 & 0.211 & 0.155 & 0.176 & 0.367 & 0.289 \\
    & 60\% & 0.468 & 0.300 & 0.214 & 0.159 & 0.196 & 0.382 & 0.308 \\
    & 75\% & 0.479 & 0.307 & 0.221 & 0.164 & 0.195 & 0.385 & 0.364 \\
    \midrule
    \multirow{1}{*}{\textbf{Dynamic}}
    & \textbf{varying} & \textbf{0.516} & \textbf{0.353} & \textbf{0.278} & \textbf{0.204} & \textbf{0.233} & \textbf{0.386} & \textbf{0.469} \\
    \bottomrule    
\end{tabular}
\end{table*}

The ablation study results presented in Table~\ref{ablation_study} demonstrate the contribution of each individual component to the overall performance of DTrace. Below, we provide a concise discussion of the impact of each component of DTrace.

\noindent \textbf{Bi-directional Multimodal Generation:} Incorporating bi-directional generation significantly improved the model's ability to capture mutual associations between medical images and their corresponding reports. This enhancement led to a notable increase in CIDEr, from 0.143 to 0.241, and a slight improvement in the F1 score to 0.280. The bi-directional nature of the approach allows the model to better align the content of the generated reports with visual features, thereby improving both the relevance and accuracy of the textual output.

\noindent \textbf{Dynamic Learning:} The introduction of dynamic learning resulted in performance gain across all metrics, particularly in BLEU-4, which increased from 0.107 to 0.120, and the F1 score, which rose to 0.344. The dynamic learning strategy compels each modality not only to reconstruct its own content but also to rely on the complementary modality for missing semantics. As a result, the model simultaneously learns stronger unimodal features and more aligned visual–language features, enhancing cross-modal fusion's effectiveness. The observed gains in NLG metrics indicate that the model is no longer confined to reproducing surface-level regularities or statistical patterns in the training corpus. Instead, it develops a stronger perception ability to capture salient pathological semantics. In parallel, the improvements in CE metrics demonstrate that the generated contents are more clinically aligned with radiological findings, reflecting that the strategy enhances linguistic fluency and ensures that the generated reports faithfully convey medically relevant information.  

\noindent\textbf{Traceback Mechanism:} The traceback mechanism provided the most substantial improvements. By supervising the semantic validity of the generated content, this component enhanced the model's ability to produce clinically accurate and coherent reports. The BLEU-4 score increased to 0.129, while CIDEr reached 0.311, indicating improved alignment with the true medical meaning of the reports. Clinically, the model achieved a precision of 0.411, a recall of 0.436, and an F1 score of 0.391. These improvements underscore the traceback mechanism's critical role in ensuring that the generated text not only follows the expected structure but also conveys accurate and meaningful clinical information. Notably, while the reconstruction-only multimodal autoencoder baseline achieves moderate results, its performance is consistently inferior to our traceback design. This shows that reconstruction alone primarily captures surface-level morphological or lexical regularities, but fails to guarantee semantic correctness. In contrast, the traceback mechanism reintegrates generated outputs into the encoders, compelling the model to verify and refine pathological semantics.

\noindent\textbf{Understanding of Cross-modal Knowledge:} The DTrace framework demonstrates clear superiority over both the baseline Encoder-Decoder model and the Multimodal Masked Autoencoder. While the baseline model struggles to generate clinically accurate reports, often relying on statistical regularities, DTrace's dynamic learning strategy and traceback mechanism ensure that the generated content is both semantically valid and clinically meaningful. Compared to the Multimodal Masked Autoencoder, which integrates vision and language modalities but faces challenges in maintaining semantic precision, DTrace excels by dynamically adapting to varying input proportions and supervising the clinical relevance of the generated content.

\begin{figure*}[t]
\centering
\includegraphics[width=0.95\textwidth]{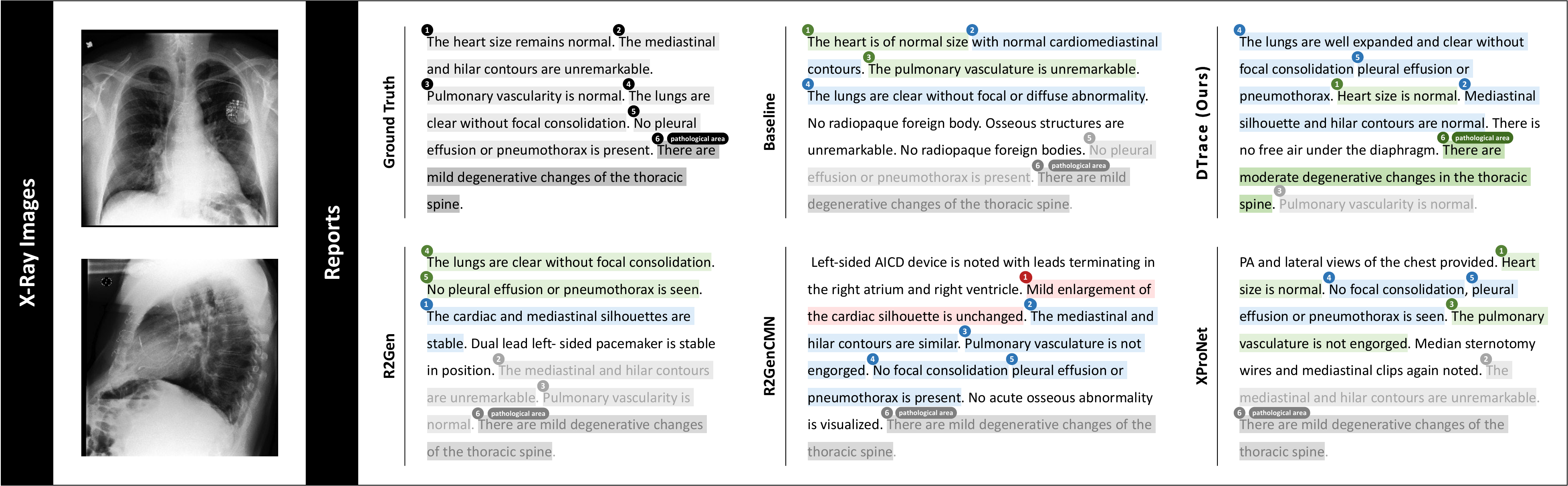} 
\caption{An example (patient 10014765) of comparisons between different report generation frameworks and the proposed DTrace framework. The information in the ground truth report is labeled from 1 to 6 and highlighted separately. The generated reports are labeled according to the ground truth report and highlighted with different colors to represent the differences between the generated sequences and the ground truth report: (1) Green - consistent; (2) Blue - semantically similar but different in expression; (3) Pink - incorrect information; (4) Gray - missing sentences; 5) Unhighlighted - not included in the ground truth.}
\label{reports_comparison}
\end{figure*}

\begin{figure*}[t]
\centering
\includegraphics[width=0.95\textwidth]{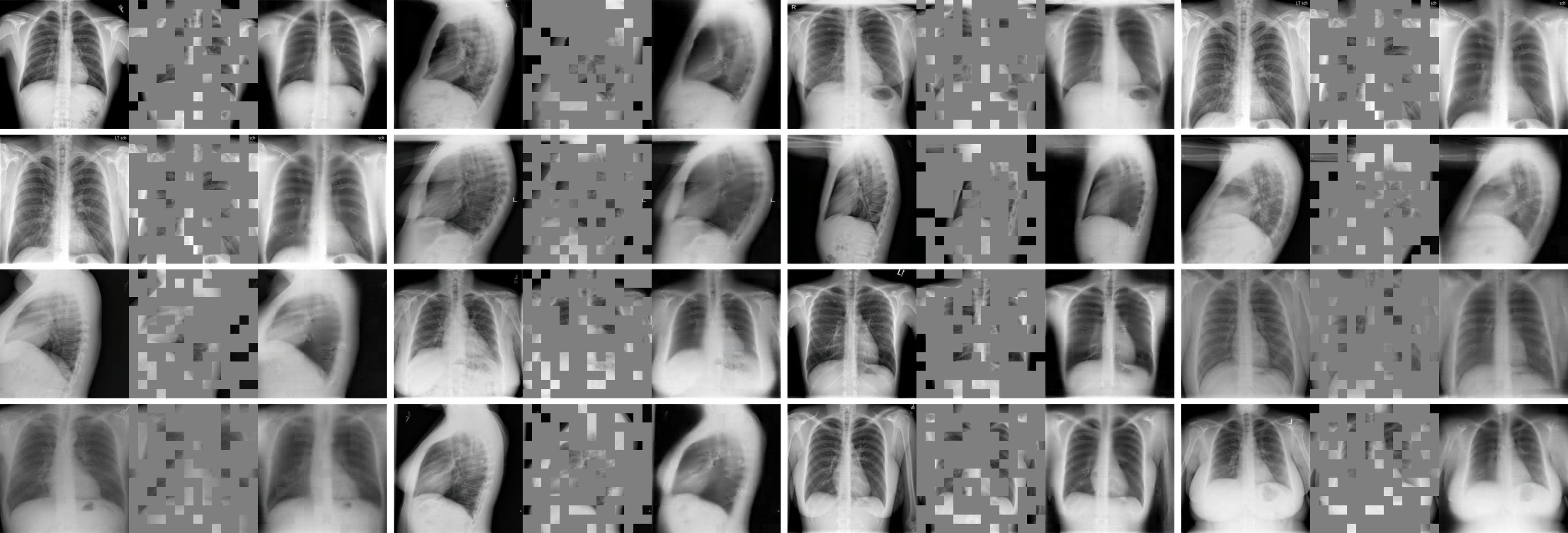} 
\caption{Visualization of the reconstructed images with a mask ratio of $75\%$ from the model. For each triplet, we show the original image (left), the masked image (middle), and our reconstructed image (right).}
\label{visualization}
\end{figure*}

\subsection{Impact of Mask Ratios}
\label{impact_of_mask_ratios}

To illustrate the significance of dynamic learning in medical report generation, we compared traditional methods that use a fixed mask ratio with our proposed method, which employs a varying mask ratio, as shown in Table~\ref{masks_comparison}. Our comparison revealed that the model’s performance improves as the mask ratio increases. Importantly, during training, the model is exposed to both image and text data, whereas during inference, it processes only the image data. The greater the discrepancy between the information available during training and inference, the worse the model’s performance tends to be. Our dynamic learning approach effectively mitigates this issue by randomly varying the proportions of information from both modalities, thus enhancing the model’s robustness during inference.

\begin{figure}[t]
\centering
\includegraphics[width=\linewidth]{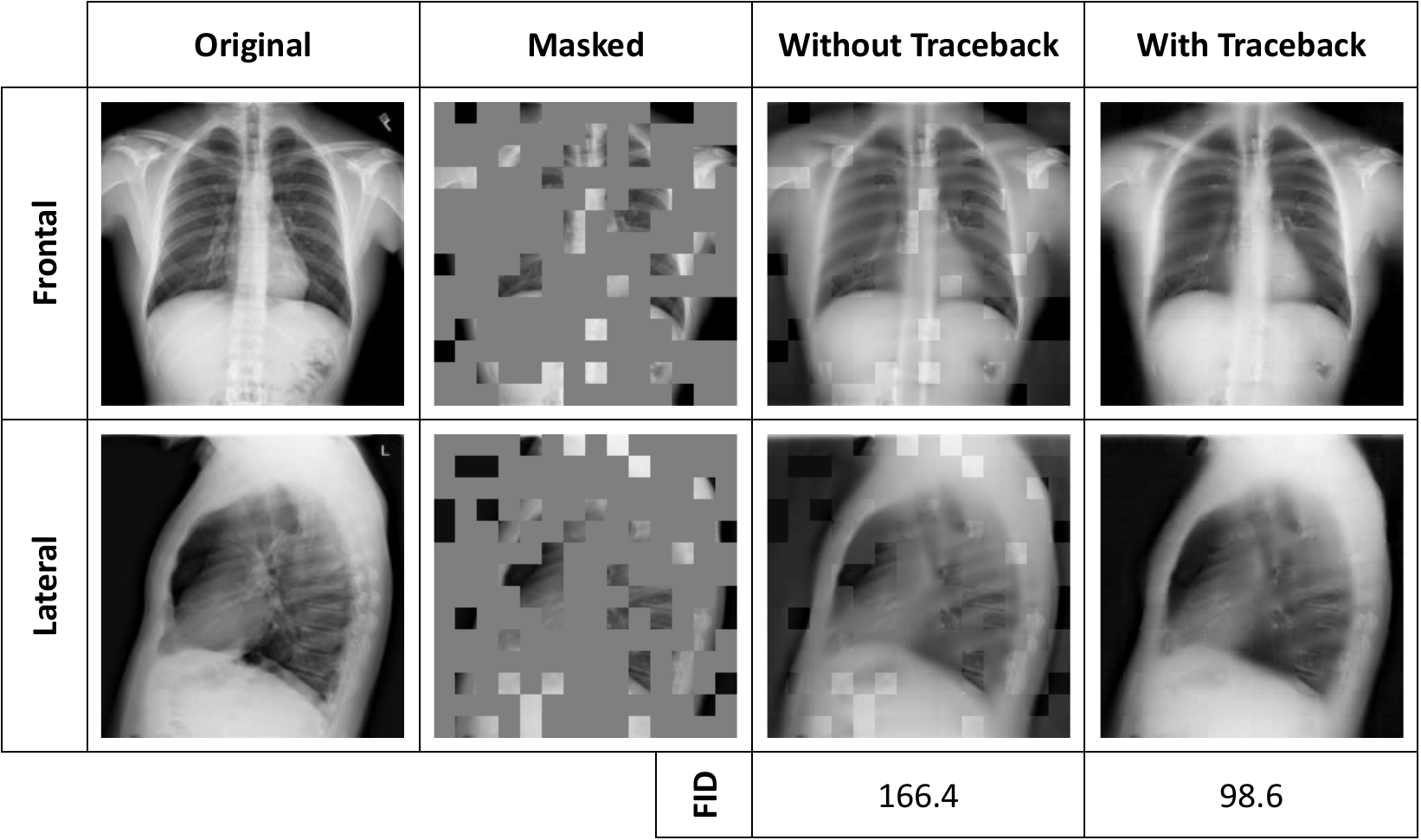} 
\caption{Comparison of visualization of reconstructed images with and without dynamic traceback learning.}
\label{fid_compare}
\end{figure}

\subsection{Qualitative Analysis and Visualization}

We conducted a qualitative analysis of DTrace compared to the baseline encoder-decoder framework. As illustrated in Fig.~\ref{reports_comparison}, when dealing with rare diseases, the baseline and existing methods frequently omit critical diagnostic statements, such as ``mild degenerative changes in the thoracic spine," despite having a high degree of textual overlap with the ground truth. In contrast, the DTrace-generated reports include most of the essential diagnostic statements and demonstrate a high level of consistency with the ground-truth reports. Additionally, Fig.~\ref{visualization} presents images reconstructed from $75\%$-masked images and unmasked reports by DTrace. These reconstructed images exhibit a high degree of consistency with the original unmasked images, highlighting DTrace's effectiveness in image reconstruction. Further visualizations can be found in the supplementary materials.

We further assessed the morphological and semantic similarities between the constructed images and the original images. To evaluate the efficacy of our dynamic traceback learning in reducing pixel-level differences, we conducted a control experiment comparing the quality of images generated with and without dynamic traceback learning, as depicted in Fig.~\ref{fid_compare}. Our analysis revealed a distinct boundary between generated patches and original patches in images not reconstructed using the dynamic traceback learning strategy. Although dynamic traceback learning is not specifically designed to enhance morphological similarity, the images reconstructed using this method appear more cohesive and clearer.

To quantify the quality of the generated images, we compared the Frechet Inception Distance (FID) scores. After implementing dynamic traceback learning, the FID of the constructed images decreased from $166.4$ to $98.6$, indicating that dynamic traceback learning more accurately mimics the distribution of the original images and produces higher-quality, more realistic reconstructions. Additionally, we employed a classification evaluation method to assess the semantic correctness of the reconstructed images.

\begin{figure*}[b]
\centering
\includegraphics[width=0.95\textwidth]{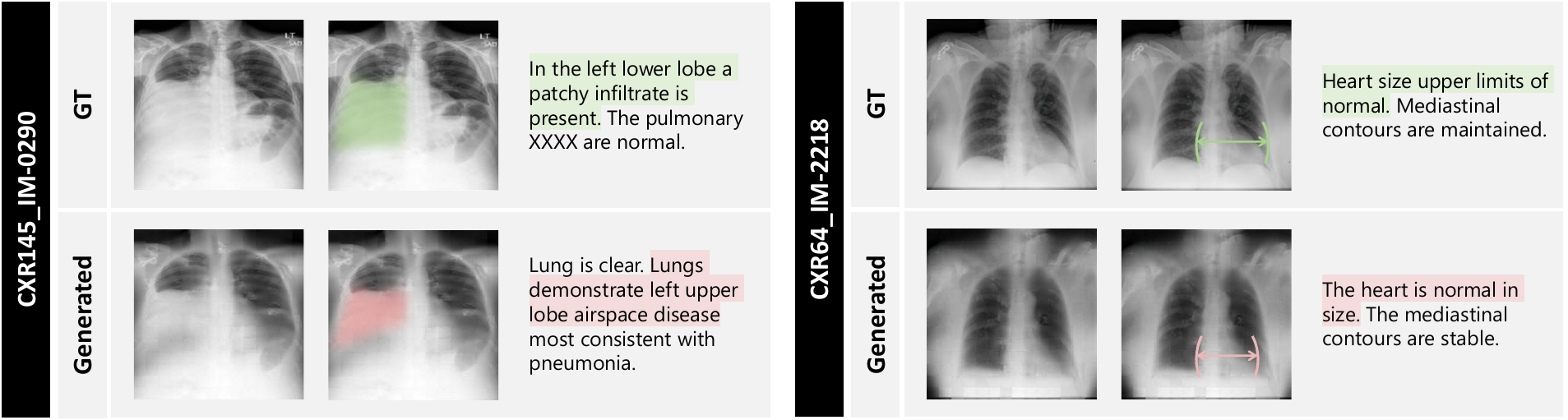} 
\caption{Interpretability analysis of DTrace. For each case, the upper row presents the ground-truth image and reference report, while the lower row shows the corresponding outputs generated by DTrace. Within each row, we display the original image (left), the highlighted region of interest (middle), and the associated report (right).}
\label{interpretability}
\end{figure*}

\subsection{Interpretability}

To better understand the association between image reconstruction and report generation, we conducted an interpretability analysis by visualizing how reconstructed images affect report generation. Figure 7 illustrates two representative cases, demonstrating how the model leverages reconstructed visual semantics to influence textual outputs. In the left case (CXR145\_IM-0290), the ground truth report states, ``In the left lower lobe, a patchy infiltrate is present." The generated report correctly describes, ``Lungs demonstrate left lobe airspace disease most consistent with pneumonia." Although the terminology differs, these descriptions are clinically consistent. In radiologic interpretation, a patchy infiltrate refers to irregular, scattered areas of increased lung opacity, suggesting partial filling of the alveoli by fluid, pus, blood, or cells, without a uniform or lobar distribution. Conversely, lobe airspace disease denotes a more localized and confluent filling of alveoli within a single lung lobe, as seen in lobar pneumonia. When patchy infiltrates are predominantly confined to one lobe and demonstrate airspace-filling features, they are synonymous with lobe airspace disease. Thus, the model’s output captures the same underlying pathology despite the lexical variation. The reconstructed image and highlighted region of interest also exhibit strong alignment with the original abnormality, indicating that the traceback mechanism preserves semantic information critical for accurate report generation. In contrast, the right case (CXR64\_IM-2218) illustrates a failure scenario. The ground truth report notes ``Heart size is mildly enlarged," yet the generated report incorrectly states, ``The heart is normal in size." Here, the reconstructed image erroneously depicts the heart as normal, leading to a semantic mismatch between visual reconstruction and textual output. This example highlights an important limitation: when the reconstruction of a key anatomical structure is inaccurate, errors can be propagated into the generated report. Collectively, these cases demonstrate that reconstructed images function as a transparent window into the model’s decision-making process, revealing both its ability to preserve nuanced semantic consistency and its vulnerability when foundational visual cues are misrepresented.

\subsection{Learning Process}
\label{learning process}

\begin{figure}[t]
\centering
\includegraphics[width=\linewidth]{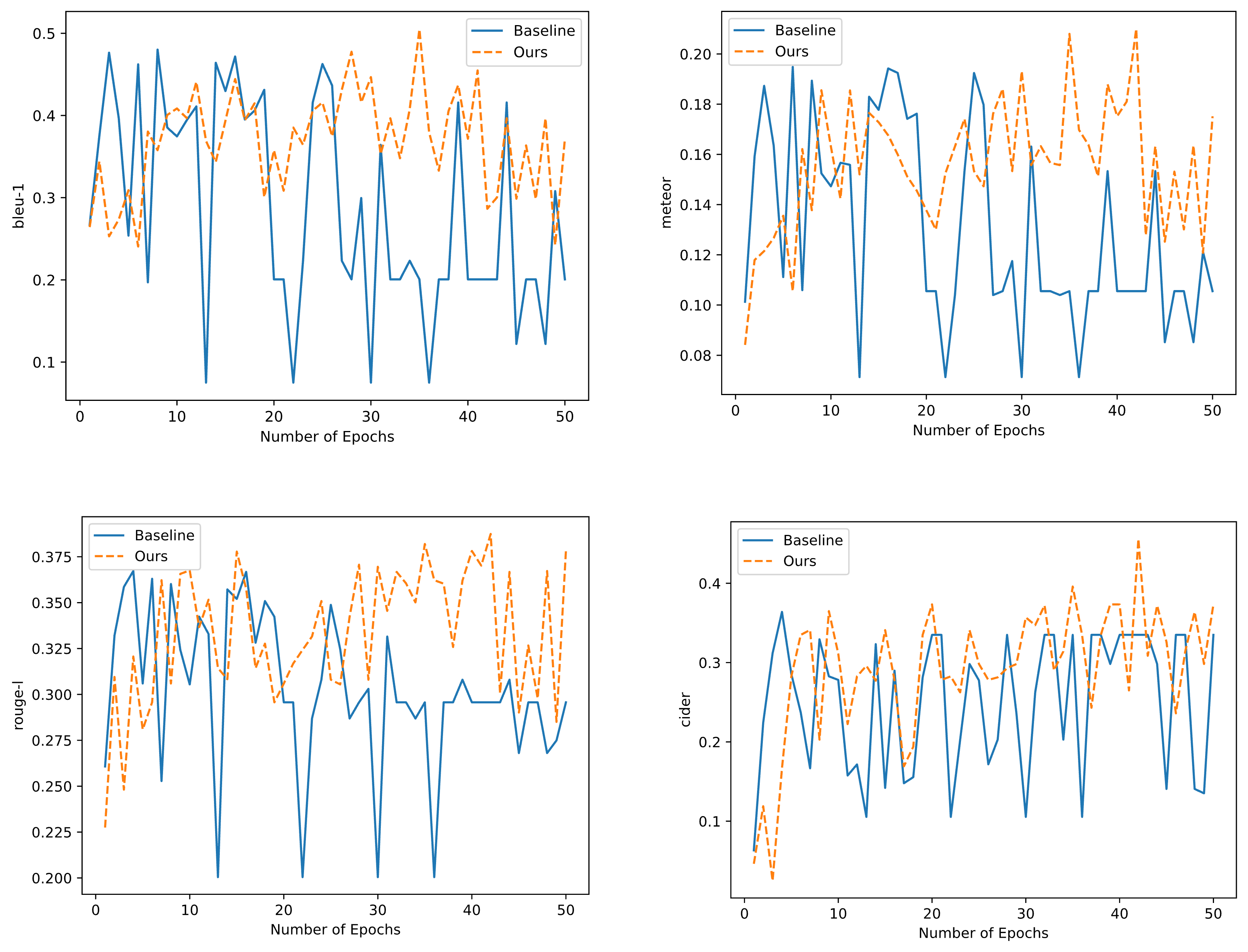} 
\caption{Comparative Visualization of Learning Process. The x-axis represents the epoch count, while the y-axis quantifies evaluation metrics (BLEU, METEOR, ROUGE-L, and CIDEr). The baseline model, featuring a traditional encoder-decoder architecture, is delineated by the blue line, whereas our novel framework is depicted by the orange line.}
\label{training}
\end{figure}

Our investigation further delved into the model's learning process by visualizing the evaluation metrics of the validation set as a function of epochs, as depicted in Fig.~\ref{training}. We noted that traditional models for report generation exhibited considerable fluctuations in performance metrics during training and were prone to entrapment in local minima. When trapped in a local minimum, the model tended to produce ``average reports," generating identical reports for any input image, which corresponded to a relatively low cross-entropy loss. However, such models are practically futile, as this is a consequence of data imbalance and inconsistency in report expression. In cases where descriptions pertain to a specific organ, the majority of instances are deemed normal, leading the model to a complacency that precludes learning how to extract features and transmit visual information to the text generator. To minimize cross-entropy loss, the model might adopt a uniform expression approach. In short, traditional methods, in our context, incline the model to learn a template that minimizes cross-entropy loss, rather than diagnosing the radiology images. By incorporating a traceback mechanism and a dynamic learning strategy, our model is compelled to engage in exchange and extract semantic information. Revealing parts of the report apprises the model of the general expression style of the input, thereby concentrating its focus on the alignment of key information. Consequently, this approach results in smaller fluctuations in performance metrics during training and enhances overall performance.

\section{Limitations}
\label{limitations}

The primary limitation of this study is the elevated computational cost incurred by the additional image decoder and text encoder during the training phase. However, once the multimodal generation model is fully trained, it becomes feasible to employ pruning techniques to remove components that are irrelevant to specific tasks, such as report generation in our scenario.

\section{Conclusion}

In this study, we introduced a novel medical report generation framework, DTrace, which leverages multimodal dynamic traceback learning. We incorporated a traceback mechanism to ensure semantic correctness during training and a dynamic learning strategy to reduce the dependency of existing generative cross-modal frameworks on textual input. Experimental evaluations demonstrate that the incorporation of these components significantly enhances the multimodal generation capabilities of the framework, enabling DTrace to achieve state-of-the-art performance on the benchmark IU-Xray and MIMIC-CXR datasets.

\bibliographystyle{IEEEtran}
\bibliography{references.bib}

\newpage

\begin{IEEEbiography}[{\includegraphics[width=1in,height=1.25in,clip,keepaspectratio]{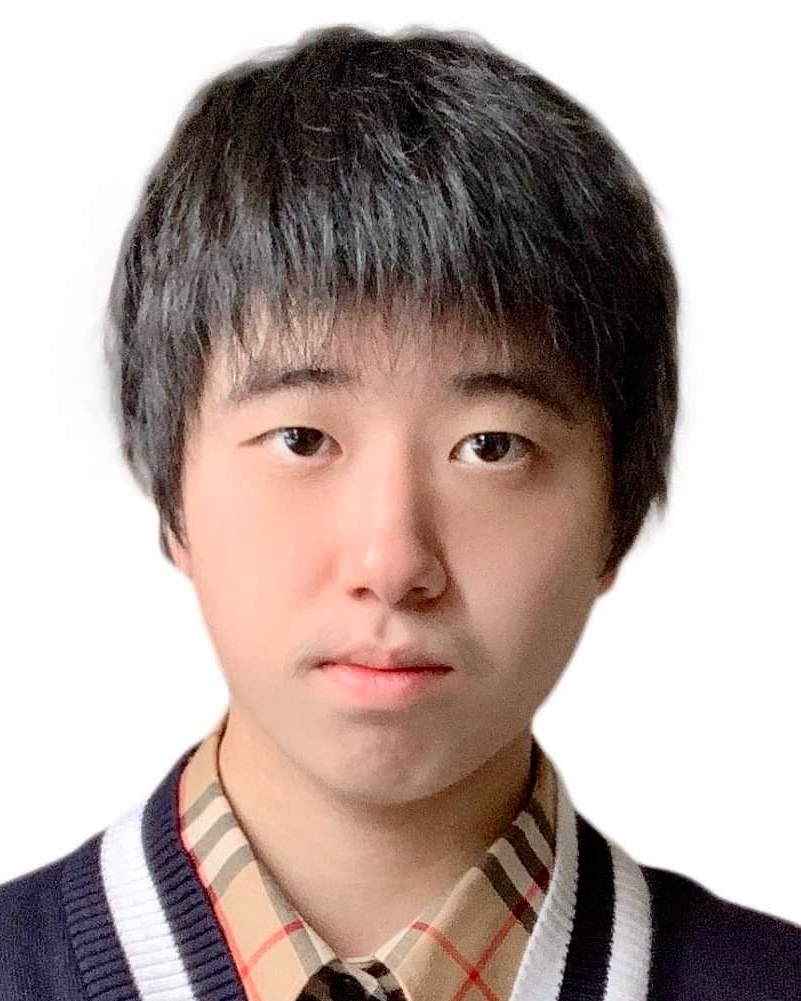}}]{Shuchang Ye}
received the Bachelor of Advanced Computing (Honours) degree in Computer Science and Computational Data Science from the University of Sydney, Australia, in 2024. He is currently pursuing the Ph.D. degree in Engineering at the University of Sydney. He was awarded the Ian Jackson Memorial Prize for Computer Science, the Dean’s List for Academic Excellence, and the Dalyell Scholarship for outstanding performance. His research interests include multimodal learning, medical image analysis and vision-language models. He serves as an active reviewer for top international conferences and journals such as MICCAI, ICCV, IEEE Transactions on Medical Imaging (TMI), and npj Digital Medicine. His research has been published in leading venues including IEEE Transactions on Multimedia, ICCV, MICCAI, and ACM MM.
\end{IEEEbiography}

\begin{IEEEbiography}[{\includegraphics[width=1in,height=1.25in,clip,keepaspectratio]{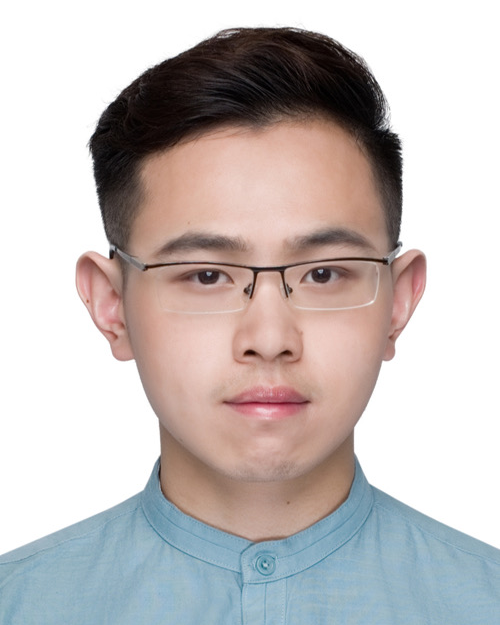}}]{Mingyuan Meng}
received his BE degree from Tsinghua University, China, in 2018 and received his MPhil/PhD degrees from the University of Sydney, Australia, in 2021/2025. He also serves as a visiting scholar at the Institute of Translational Medicine, Shanghai Jiao Tong University, China. His current research focus is deep learning for medical image analysis, especially in medical image dense prediction, radiomics, and vision-language modeling.
\end{IEEEbiography}

\begin{IEEEbiography}[{\includegraphics[width=1in,height=1.25in,clip,keepaspectratio]{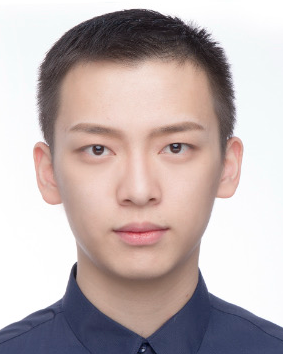}}]{Mingjian Li}
received the B.Sc. degree in Biomedical Engineering from Huazhong University of Science and Technology, Wuhan, China, in 2016, the M.Sc. degree from Shanghai Jiao Tong University, Shanghai, China, in 2019, and the Ph.D. degree from the University of Sydney, Sydney, NSW, Australia, in 2024. His current research interests include medical image visualization and vision–language pretraining.
\end{IEEEbiography}

\begin{IEEEbiography}[{\includegraphics[width=1in,height=1.25in,clip,keepaspectratio]{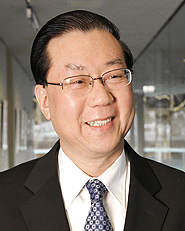}}]{Dagan Feng}
is Director, Biomedical \& Multimedia Information Technology (BMIT) Research Group, and Funding Head, School of Information Technology (renamed as Computer Science recently) and Funding Director, Institute of Biomedical Engineering and Technology Institute, the University of Sydney. He received his Master of Engineering in Electrical Engineering \& Computer Science (EECS) from Shanghai Jiao Tong University in 1982, Master of Science in Biocybernetics and PhD in Computer Science from the University of California, Los Angeles (UCLA) in 1985 and 1988 respectively, where he received the Crump Prize for Excellence in Medical Engineering. In conjunction with his team members and students, he has been responsible for more than 50 key research projects, published over 1000 scholarly research papers, pioneered several new research directions, and made a number of landmark contributions in his field. He has served as Chair of the International Federation of Automatic Control (IFAC) Technical Committee on Biological and Medical Systems, Special Area Editor / Associate Editor / Editorial Board Member for a dozen of core journals in his area, and Scientific Advisor for a couple of prestigious organizations. He has been invited to give over 100 keynote presentations in 23 countries and regions, and has organized / chaired over 100 major international conferences / symposia / workshops. Professor Feng is Fellow of ACS, HKIE, IET, IEEE, and Australian Academy of Technological Sciences and Engineering.
\end{IEEEbiography}

\begin{IEEEbiography}[{\includegraphics[width=1in,height=1.25in,clip,keepaspectratio]{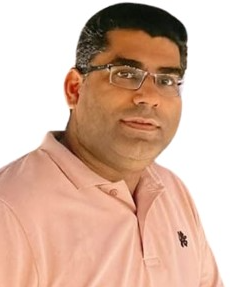}}]{Usman Naseem}
is an Assistant Professor at the School of Computing, Macquarie University, Australia. His research focuses on natural language processing (NLP), with emphasis on aligning large language models (LLMs) with human values and preferences, trust and safety, and NLP for social good. He has authored over 100 publications in venues such as ACL, EMNLP, NAACL, AAAI, and WebConf, and his work has received multiple Best Paper Awards. He has served as Senior Area Chair for EMNLP 2025, Co-Chair for Web4Good at WebConf 2026, and Area Chair or Senior Program Committee member for leading *CL and AI conferences including ACL, EMNLP, AAAI, and EACL. He has also organized multiple shared tasks, including SemEval 2026, as well as top-tier workshops such as MM4SG (WebConf 2024–2025, ICDM 2024) and VLM4Bio (ACM MM 2024).
\end{IEEEbiography}

\begin{IEEEbiography}[{\includegraphics[width=1in,height=1.25in,clip,keepaspectratio]{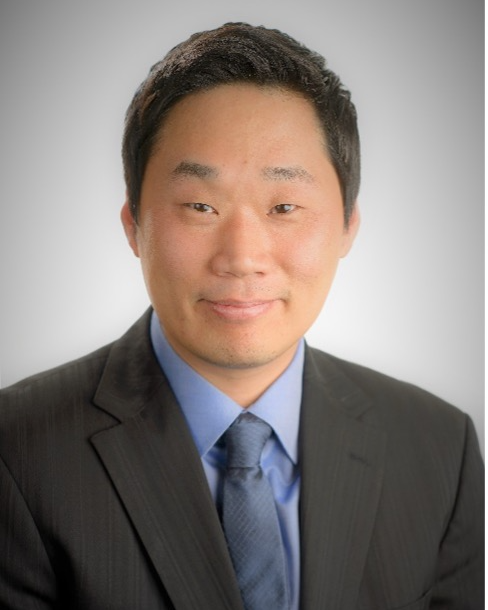}}]{Jinman Kim}
is a Professor of Computer Science at the University of Sydney. He received his PhD from the University of Sydney in 2006 and was an Australian Research Council (ARC) Postdoctoral Research Fellow at the University of Sydney and then a Marie Curie Senior Research Fellow at the University of Geneva prior to joining the University of Sydney in 2013 as a faculty member. He is currently an ARC industry fellow, closely collaborating with his industry partner, Royal Prince Alfred Hospital, to conduct translational research. Prof Kim’s research is on the application of machine learning to biomedical image analysis and visualization. His focus is on cross-model and multimodal learning, which includes biomedical visual-language representations, image-omics, multimodal data processing, and biomedical mixed reality technologies. He established and leads the Biomedical Data Analysis and Visualisation (BDAV) Lab at the School of Computer Science. 
\end{IEEEbiography}

\end{document}